\documentclass[10pt,sigconf,screen=true,urlbreakonhyphens=false]{acmart}

\copyrightyear{2026}
\acmYear{2026}
\setcopyright{cc}
\setcctype{by}
\acmConference[ICCAD '26]{IEEE/ACM International Conference on Computer-Aided Design}{November 08--12, 2026}{San Jose, CA, USA}
\acmBooktitle{IEEE/ACM International Conference on Computer-Aided Design (ICCAD '26), November 08--12, 2026, San Jose, CA, USA}
\acmDOI{10.1145/3831252.3834253}
\acmISBN{979-8-4007-2873-0/2026/11}

\usepackage{graphicx}
\usepackage{booktabs}
\usepackage{multirow}
\usepackage{amsmath}
\usepackage{algorithm}
\usepackage{algorithmic}
\usepackage{pifont}
\usepackage{xcolor}
\usepackage{colortbl}
\usepackage{enumitem}
\usepackage{threeparttable}
\usepackage[most]{tcolorbox}
\usepackage{cleveref}
\usepackage{verbatim}

\hypersetup{
  colorlinks=true,
  linkcolor=blue,
  citecolor=blue,
  urlcolor=blue,
}

\usepackage{xpatch}
\makeatletter
\xpatchcmd{\paragraph}{3.25ex \@plus 1ex \@minus .2ex}{0.6ex \@plus 0.3ex \@minus 0.2ex}{}{}
\makeatother

\Crefname{figure}{Fig.}{Figs.}
\Crefname{table}{Table}{Tables}
\Crefname{section}{Sec.}{Secs.}
\Crefname{equation}{Eq.}{Eqs.}

\definecolor{myblue}{RGB}{225,235,246}
\definecolor{mygreen}{RGB}{202,223,184}
\definecolor{myorange}{RGB}{244,106,18}

\definecolor{sysblue}  {HTML}{1B4F72}
\definecolor{sysbg}    {HTML}{D6EAF8}
\definecolor{userbrown}{HTML}{6E2C00}
\definecolor{userbg}   {HTML}{FDEBD0}
\definecolor{codegray} {HTML}{F2F3F4}
\definecolor{hdrbg}    {HTML}{17202A}

\tcbset{
  sysprompt/.style = {
    enhanced, colback=sysbg, colframe=sysblue,
    boxrule=0.8pt, arc=3pt,
    title={\small\bfseries System Prompt},
    coltitle=white, colbacktitle=sysblue,
    top=4pt, bottom=4pt, left=6pt, right=6pt,
    fontupper=\small,
  },
  userprompt/.style = {
    enhanced, colback=userbg, colframe=userbrown,
    boxrule=0.8pt, arc=3pt,
    title={\small\bfseries User Prompt},
    coltitle=white, colbacktitle=userbrown,
    top=4pt, bottom=4pt, left=6pt, right=6pt,
    fontupper=\small,
  },
  stagehr/.style = {
    enhanced, colback=hdrbg, colframe=hdrbg,
    arc=3pt, boxrule=0pt,
    top=3pt, bottom=3pt, left=6pt, right=6pt,
    halign=center,
    fontupper=\small\bfseries\color{white},
  },
}

\usepackage{balance}   
\graphicspath{{./figures/}}

\begin{document}

\title{AnalogRetriever: Learning Cross-Modal Representations for Analog Circuit Retrieval}

\author{Yihan Wang}
\authornote{Both authors contributed equally to this research.}
\affiliation{\institution{Tsinghua University}\city{Beijing}\country{China}}
\email{yihan-wa23@mails.tsinghua.edu.cn}

\author{Lei Li}
\authornotemark[1]
\affiliation{\institution{The University of Hong Kong}\city{Hong Kong}\country{China}}
\email{nlp.lilei@gmail.com}

\author{Yao Lai}
\affiliation{\institution{University of Cambridge}\city{Cambridge}\country{United Kingdom}}
\email{yl2204@cam.ac.uk}

\author{Jing Wang}
\authornote{Corresponding authors.}
\affiliation{\institution{Nanjing University of Posts and Telecommunications}\city{Nanjing}\country{China}}
\email{wangjing25@njupt.edu.cn}

\author{Yan Lu}
\authornotemark[2]
\affiliation{\institution{Tsinghua University}\city{Beijing}\country{China}}
\email{yanlu@tsinghua.edu.cn}

\renewcommand{\shortauthors}{Wang et al.}

\begin{abstract}
Analog circuit design relies heavily on reusing existing intellectual property (IP), yet searching across heterogeneous representations such as SPICE netlists, schematics, and functional descriptions remains challenging. Existing methods are largely limited to exact matching within a single modality, failing to capture cross-modal semantic relationships. To bridge this gap, we present AnalogRetriever, a unified tri-modal retrieval framework for analog circuit search. We first build a high-quality dataset on top of Masala-CHAI through a two-stage repair pipeline that raises the netlist compile rate from 22.0\% to 99.97\%. Built on this foundation, AnalogRetriever encodes schematics and descriptions with a vision-language model and netlists with a port-aware relational graph convolutional network, mapping all three modalities into a shared embedding space via curriculum contrastive learning. Experiments show that AnalogRetriever achieves an average Recall@1 of 75.2\% across all six cross-modal retrieval directions, significantly outperforming existing baselines. When integrated into the AnalogCoder agentic framework as a retrieval-augmented generation module, it consistently improves functional pass rates and enables previously unsolved tasks to be completed.
Code and data: \url{https://github.com/yihan-wa/AnalogRetriever_ICCAD}.
\end{abstract}



\maketitle

\section{Introduction}
Recent advances in large language models (LLMs) have opened new
opportunities for analog circuit design automation. Existing efforts
focus on \emph{generative} approaches that synthesize designs from
specifications or generate
netlists from schematic images,
but they suffer from hallucination, invalid topologies, and difficulty
incorporating domain-specific constraints.

\begin{figure}[!t]
    \centering
    \makebox[\columnwidth][c]{\includegraphics[width=1\columnwidth]{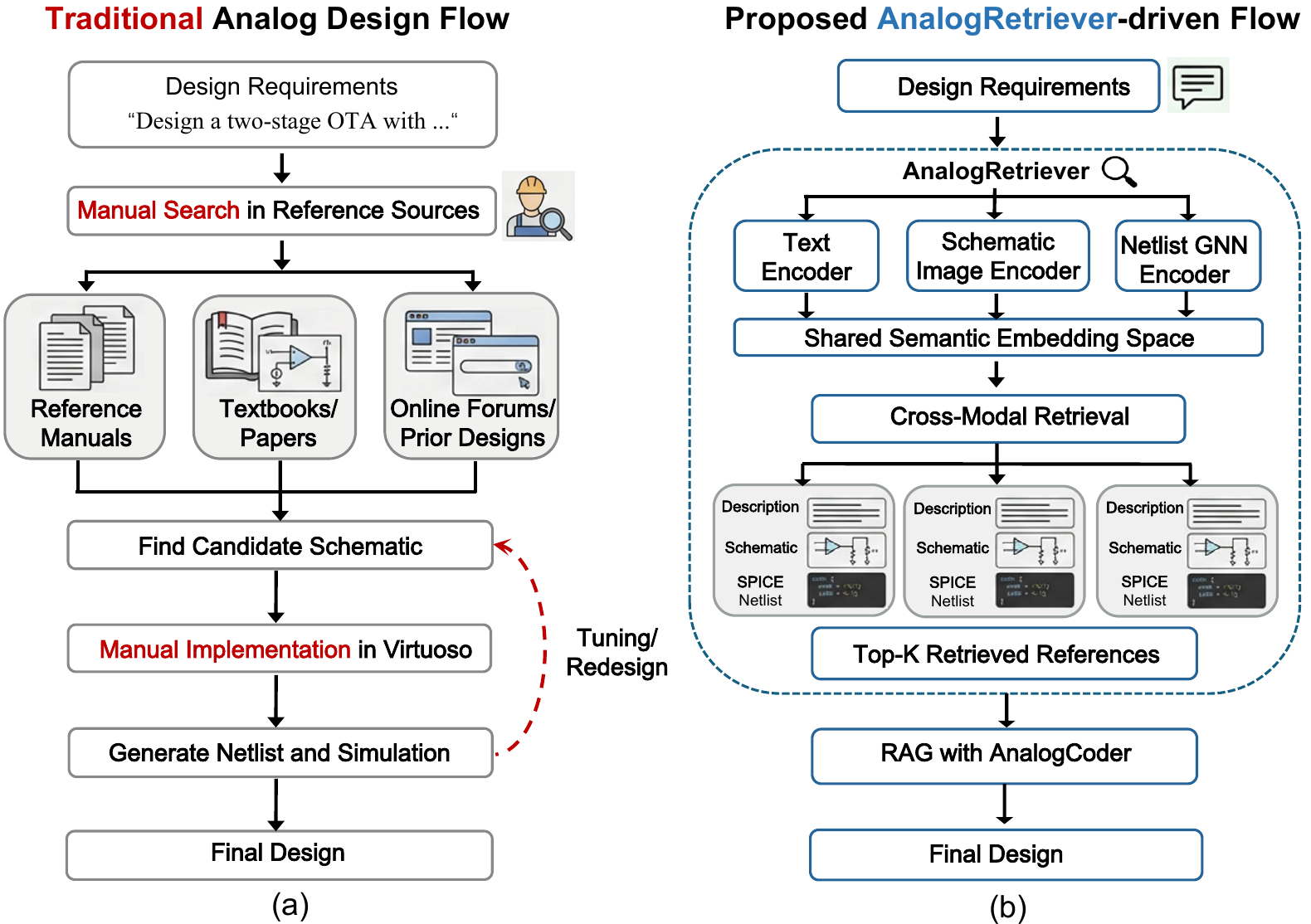}}
    \caption{Comparison of analog design flows: \textbf{(a)} Traditional flow based on manual search. \textbf{(b)} Proposed AnalogRetriever-driven flow based on cross-modal retrieval and retrieval-augmented generation.}
    \label{fig:motivation}
\end{figure}

In contrast, \emph{retrieval-based} approaches for analog design remain
largely unexplored despite their practical potential. As shown in
\Cref{fig:motivation}(a), junior engineers today spend considerable
time manually searching design manuals, papers, and internal repositories
with keyword queries, a process that is time-consuming, expertise-heavy,
and especially hard for newcomers who may not know the right terminology.
The challenge is compounded by the heterogeneous nature of the circuit
representations: the same circuit exists as a SPICE netlist (code), a schematic (image), or a functional description (text), yet existing circuit retrieval systems operate on a single modality~\cite{11240643}.

Unlike digital design, which benefits from mature synthesis and
automation flows, analog design relies heavily on reusing proven
topologies accumulated through years of engineering
experience~\cite{Chen_2024}. 
An effective cross-modal retrieval system would let engineers describe requirements in natural language and obtain matching schematics and netlists (\Cref{fig:motivation}(b)), accelerating design exploration, facilitating intellectual property (IP) reuse, and lowering the entry barrier for junior designers. 
Such a system also complements generative methods through
retrieval-augmented generation (RAG), where verified existing designs
ground LLM outputs and mitigate hallucination. However, building it
requires bridging the representational gap between netlists,
schematics, and functional
descriptions, a cross-modal alignment challenge that remains
open.

To address this gap, we propose \textbf{AnalogRetriever}, a tri-modal
retrieval framework that maps functional descriptions, schematics, and
SPICE netlists into a unified semantic space via contrastive
learning~\cite{oord2019representationlearningcontrastivepredictive}. This enables flexible cross-modal retrieval:
a natural-language query such as ``a two-stage op-amp with Miller
compensation'' returns matching schematics and netlists, and a
schematic or netlist query returns functionally similar designs with
their specifications.

Building this framework requires addressing two key challenges that
no existing method tackles jointly.

\textbf{(Challenge~1)~Domain gap for vision-language alignment.}
Pretrained vision-language models (VLMs) such as
CLIP~\cite{pmlr-v139-radford21a} excel at aligning natural photographs with
captions but do not generalize to abstract circuit schematics: their
zero-shot average Recall@1 (R@1) is near random ($2.5\%$,
Table~\ref{tab:ablation}). Circuit schematics are clean line drawings
with domain-specific symbols outside CLIP's pretraining distribution,
and the largest public dataset (Masala-CHAI) suffers from severe
quality issues: only 22\% of netlists compile, hindering effective
fine-tuning.

\textbf{(Challenge~2)~Graph-structured netlists and fine-grained discrimination.}
SPICE netlists are fundamentally graph-structured: node names are
arbitrary, topology is implicit in connectivity, and standard text
encoders cannot capture these structural semantics. Moreover, circuits
in the same functional category (e.g., different op-amp topologies)
share similar descriptions but differ substantially in implementation,
making it easy for a contrastive model to conflate them.

To tackle Challenge~1, we refine Masala-CHAI through a two-stage LLM-based
repair pipeline and freeze the lower CLIP layers for domain
adaptation without catastrophic forgetting. To tackle Challenge~2, we employ a
port-aware Relational Graph Convolutional Network (RGCN) encoder with
curriculum-guided hard-negative mining.

We evaluate AnalogRetriever on a curated
Masala-CHAI~\cite{Bhandari2024MasalaCHAIAL}, where our two-stage
repair pipeline yields 6{,}371 refined triplets with a 99.97\%
compilation rate. AnalogRetriever achieves an average R@1 of $75.2\%$
across all six cross-modal directions, outperforming the strongest
zero-shot baseline (CROP~\cite{11240643}, $4.7\%$) by over $15{\times}$,
and the tri-modal training also improves the bi-modal image--text
directions. When integrated with AnalogCoder via RAG, it lifts
functional correctness on \emph{all eight} evaluated LLMs (averaging
$+5.6\%$) and sets a new state of the art of $86.7\%$ on
Claude~Sonnet~4.6.

Our main contributions are:
\textbf{(1)~Tri-Modal Topology-Aware Retrieval.} We propose
AnalogRetriever, which combines a pretrained VLM with a
topology-aware graph neural network (GNN) encoder through
tri-modal contrastive learning to enable the first cross-modal
search across text, schematics, and SPICE netlists.
\textbf{(2)~Curriculum-Guided Hard Negative Mining.} A training
strategy that progressively increases intra-cluster negatives
based on functional category, sharpening discrimination among
structurally distinct circuits with similar functionality.
\textbf{(3)~High-Quality Tri-Modal Dataset.} A two-stage LLM-based
refinement pipeline that audits and repairs Masala-CHAI, producing
6{,}371 simulator-checked triplets that we will \mbox{release publicly.}

\section{Related Work}
Prior work relevant to AnalogRetriever spans three areas: circuit
dataset construction and schematic-netlist conversion, LLM-based
analog circuit design, and cross-modal retrieval with contrastive
learning. No existing work addresses all three modalities within a
single retrieval framework.

\subsection{Schematic-Netlist Conversion}
\label{sec:rw_data}
AMSNet~\cite{10691781} and AMSNet 2.0~\cite{11105794} pioneered automatic
schematic-to-netlist conversion.
Masala-CHAI~\cite{Bhandari2024MasalaCHAIAL} scaled dataset creation via
end-to-end SPICE generation from schematics, and
Image2Net~\cite{11100581} contributed more diverse pairs.
Wang et al.~\cite{10.1145/3489517.3530410} showed that topology-aware
encoders capture circuit semantics beyond structural similarity, and
Netlistify~\cite{10.1109/MLCAD65511.2025.11189145} tackled deterministic
schematic-to-netlist conversion through component recognition. Our
framework extends these topology-aware representations to align
netlists with schematics and text via contrastive learning.

\subsection{LLMs for Analog Circuit Design}
\label{sec:rw_llm}
AnalogCoder~\cite{Lai_2025} introduced the first
training-free LLM agent for analog design via Python code generation;
AnalogCoder-Pro~\cite{11432899} extended it with multimodal
topology synthesis, and LLMs have likewise been applied to register-transfer level (RTL) code
optimization under IP constraints~\cite{wang2025principleguidedverilogoptimizationipsafe}.
AnalogXpert~\cite{11100627} formulates topology synthesis
as subcircuit-level SPICE generation,
Chen et al.~\cite{chen2026datasetconstructiontrainingllm} curate a
textbook-derived corpus to train domain-specific LLMs, LaMAGIC~\cite{pmlr-v235-chang24c} casts topology
generation as language modelling, Artisan~\cite{10.1145/3649329.3655903}
automates op-amp design end-to-end, and
AnalogGenie~\cite{ICLR2025_f0552f14} explicitly explores the topology
space. These methods focus on circuit \emph{generation} but suffer from
hallucination and invalid topologies. Our retrieval component grounds
LLM-based tools with relevant circuit examples via RAG, reducing
invalid outputs.

\subsection{Cross-Modal Retrieval and Contrastive Learning}
\label{sec:rw_retrieval}
Contrastive vision-language pretraining~\cite{pmlr-v139-radford21a,oord2019representationlearningcontrastivepredictive}
has become the de facto approach for aligning heterogeneous modalities,
but its direct application to circuit data is fundamentally limited:
SPICE netlists are graph-structured with arbitrary node names, and
circuits sharing similar descriptions can differ substantially at the
device level. GNNs for
electronic design automation (EDA)~\cite{10.1145/3489517.3530410,kipf2017semisupervised,10.1007/978-3-319-93417-4_38} recover structural semantics
from netlists but are typically trained in isolation, without
alignment to natural language or schematics. To our knowledge, no
prior work establishes a single representation space that bridges all
three analog circuit modalities; \textsc{AnalogRetriever} closes this
gap and supports all six cross-modal directions with a unified
training objective.
\section{Method}
\begin{figure}[t]
    \centering
    \makebox[\columnwidth][c]{\includegraphics[width=1\columnwidth]{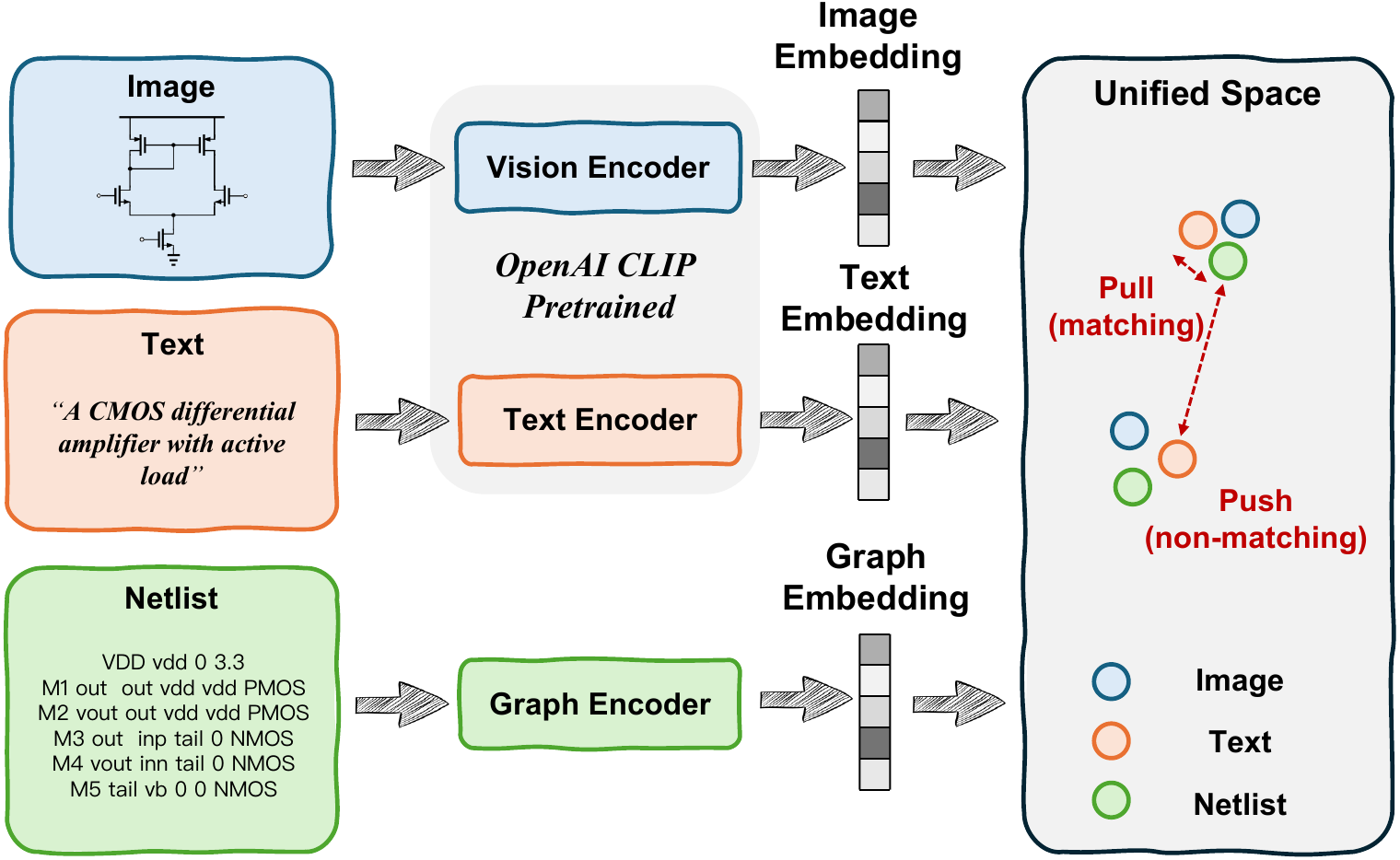}}
    \caption{Overview of AnalogRetriever: three modality-specific encoders map images, text, and netlists into a unified embedding space.}
    \label{fig:architecture}
\end{figure}

\Cref{fig:architecture} overviews the AnalogRetriever framework.
We formalize the problem, describe the modality-specific encoders,
present the tri-modal contrastive objective, and detail the
three-phase curriculum training.

\subsection{Problem Formulation}
\label{sec:formulation}
We formulate analog circuit retrieval as a tri-modal matching problem
across SPICE netlist code $\mathcal{C}$, functional text descriptions
$\mathcal{T}$, and schematic images $\mathcal{I}$. Given a dataset
$\mathcal{D} = \{(c_i, t_i, s_i)\}_{i=1}^N$ of $N$ aligned triplets
(code, text, schematic image), we learn three encoders
$f_{\mathcal{C}}, f_{\mathcal{T}}, f_{\mathcal{I}}$ that map each
modality into a shared $d$-dimensional embedding space $\mathbb{R}^d$
where semantically related items cluster together. At inference, a
query from any modality retrieves top-$K$ items from any target
modality via cosine similarity, covering all six cross-modal
directions (C$\leftrightarrow$I, C$\leftrightarrow$T,
I$\leftrightarrow$T).

\subsection{Tri-Modal Encoding Architecture}
\label{sec:encoding}

\paragraph{Vision-Language Encoding with CLIP}
We encode schematic images and textual descriptions with the pretrained
CLIP~\cite{pmlr-v139-radford21a} VLM: a Vision Transformer (ViT)-L/14~\cite{dosovitskiy2021an}
image encoder $f_{\mathcal{I}}$ and a Transformer~\cite{NIPS2017_3f5ee243}
text encoder $f_{\mathcal{T}}$, both projecting into a shared
$d{=}768$ space. To preserve the pretrained cross-modal alignment while
allowing domain adaptation to clean circuit line drawings, we freeze
the bottom 16 of 24 ViT blocks and fine-tune only the top 8, avoiding
catastrophic forgetting of CLIP's general visual-semantic prior.

\paragraph{Port-Aware Relational Graph Netlist Encoding.}
SPICE netlists are graph-structured. We parse each netlist into a graph whose nodes are both devices and electrical nets: a device is linked to every net it touches by an edge typed with the corresponding terminal, and devices sharing a net are additionally linked by a \textit{shared\_net} edge. The net nodes make connectivity explicit, while the \textit{shared\_net} edges expose device-level adjacency directly, so $L$ message-passing layers reach $L$ device hops.
Crucially, different terminals on the same device carry distinct electrical semantics: connecting to the gate versus the drain of a MOSFET implies a control input rather than a current path. 
Standard Graph Convolutional Networks (GCNs)~\cite{kipf2017semisupervised} with a single shared edge weight cannot distinguish these port-level differences, confusing functionally distinct circuits that share the same topology.
We therefore adopt an RGCN~\cite{10.1007/978-3-319-93417-4_38} with a separate learnable weight matrix per edge type. As shown in~\Cref{sec:main_results}, replacing GCN with RGCN brings the largest gains on code-involved retrieval directions, precisely where port-level semantics matter most.

Each netlist $c_i$ is parsed into a graph
$G_i{=}(\mathcal{V}_i, \mathcal{E}_i, \mathcal{R})$ with
$|\mathcal{R}|{=}20$ edge types: 19 port types covering all device
terminals---MOSFET (drain/gate/source/bulk), BJT
(collector/base/emitter), source $(\pm)$, diode (anode/cathode),
passive R/C/L terminals, four controlled-source ports,
\textit{shared\_net} for device--device connections on the same
electrical net, and \textit{subckt\_terminal} for subcircuit
interfaces---plus one reserved type for unrecognised ports. This
port-level vocabulary lets the
encoder distinguish, e.g., a MOSFET whose drain feeds the next stage
from one whose source does, which is essential for telling a common-source
stage apart from a source follower. Each device node's feature fuses a
discrete component-type embedding with a log-normalized continuous
parameter vector (channel width and length, resistance, capacitance);
net nodes use a net-type embedding (generic, supply, or ground) with a
zero parameter vector:
\begin{equation}
    \mathbf{h}_v^{(0)} = \mathbf{W}_{\text{fuse}} \big[ \text{Emb}(x_{\text{type}}) \,\|\, \text{Linear}(\log(1{+}x_{\text{cont}})) \big]
\end{equation}
where $\mathbf{W}_{\text{fuse}}$ is a learnable fusion matrix,
$\text{Emb}(\cdot)$ is a discrete embedding lookup, $x_{\text{type}}$
is the component type, $x_{\text{cont}}$ is the continuous parameter
vector, and $\|$ denotes concatenation.

Port-aware message passing at layer $l$ is
\begin{equation}
    \mathbf{h}_v^{(l+1)} = \sigma\!\Bigg(\sum_{r \in \mathcal{R}} \sum_{u \in \mathcal{N}_r(v)} \frac{1}{|\mathcal{N}_r(v)|} \mathbf{W}_r^{(l)} \mathbf{h}_u^{(l)}\Bigg) + \mathbf{h}_v^{(l)},
\end{equation}
where $\mathcal{N}_r(v)$ is the set of neighbors of node $v$ under
relation type $r$, $\mathbf{W}_r^{(l)}$ is the relation-specific weight
matrix at layer $l$, and $\sigma$ denotes the GELU activation. Each
layer further applies GraphNorm~\cite{pmlr-v139-cai21e} and a residual
connection to stabilize gradient flow. We use $L{=}2$
layers: a two-hop receptive field is sufficient to capture the
canonical subcircuit patterns that dominate analog topologies
(differential pairs, current mirrors, cascode stacks typically span
one or two device~hops).

A learnable attention pool then aggregates node embeddings into a
graph-level vector, letting the model up-weight functionally critical
nodes---input pairs, output stages, bias references, and the nets
joining them---over boilerplate:
\begin{equation}
    \mathbf{g}_i = \sum_{v \in \mathcal{V}_i} \alpha_v \mathbf{h}_v^{(L)}, \quad
    \alpha_v \propto \exp(\mathbf{w}^\top \mathbf{h}_v^{(L)}),
\end{equation}
where $\mathbf{w}$ is a learnable attention vector. Compared with
uniform sum or mean pooling, attention pooling gives the graph
embedding a clearer notion of \emph{which parts of the circuit matter}
for a given functional role. Finally, a two-layer multi-layer perceptron
(MLP) projection head
 with LayerNorm, GELU activation, and dropout
($p{=}0.1$) maps $\mathbf{g}_i$ into the CLIP space,
$\mathbf{v}_i^{(c)}{=}f_{\mathcal{C}}(c_i){=}\text{MLP}_{\text{proj}}(\mathbf{g}_i)$,
followed by $\ell_2$ normalization. The nonlinear projection provides
extra capacity to bridge the representation gap between the graph and
vision-language domains without perturbing the pretrained CLIP space.

\subsection{Tri-Modal Contrastive Learning}
\label{sec:contrastive}
We align the three modalities with an InfoNCE-style~\cite{oord2019representationlearningcontrastivepredictive}
objective. Let $\mathbf{v}_i^{(c)}, \mathbf{v}_i^{(s)}, \mathbf{v}_i^{(t)}$
denote the $\ell_2$-normalized embeddings of the $i$-th code, schematic
image, and text sample, respectively. For a batch of $B$ triplets, the
directional Code$\rightarrow$Image loss is
\begin{equation}
    \mathcal{L}_{\mathcal{C} \to \mathcal{I}} = -\frac{1}{B} \sum_{i=1}^B \log \frac{\exp(\text{sim}(\mathbf{v}_i^{(c)}, \mathbf{v}_i^{(s)}) / \tau)}{\sum_{j=1}^B \exp(\text{sim}(\mathbf{v}_i^{(c)}, \mathbf{v}_j^{(s)}) / \tau)}
\end{equation}
where $\text{sim}(\cdot,\cdot)$ denotes cosine similarity and $\tau$ is
a learnable temperature. The full tri-modal loss sums all
six directions across the three modality pairs:
\begin{equation}
\mathcal{L}_{\text{tri}} = \!\!\!\sum_{(a,b)\in\mathcal{P}}\!\! \big(\mathcal{L}_{a\to b}+\mathcal{L}_{b\to a}\big),\;\;
\mathcal{P}{=}\{(\mathcal{C},\mathcal{I}),(\mathcal{T},\mathcal{I}),(\mathcal{C},\mathcal{T})\}.
\end{equation}
We apply label smoothing $\epsilon{=}0.1$, which we find especially
helpful when training with intra-cluster hard negatives. Importantly,
the complete tri-modal training procedure produces \emph{mutual
enhancement}: the originally bi-modal Image$\leftrightarrow$Text
directions also improve, so aligning the three
modalities benefits all of them.

\subsection{Three-Phase Curriculum Training}
\label{sec:curriculum}
\begin{figure}[t]
    \centering
    \makebox[\columnwidth][c]{\includegraphics[width=0.95\columnwidth]{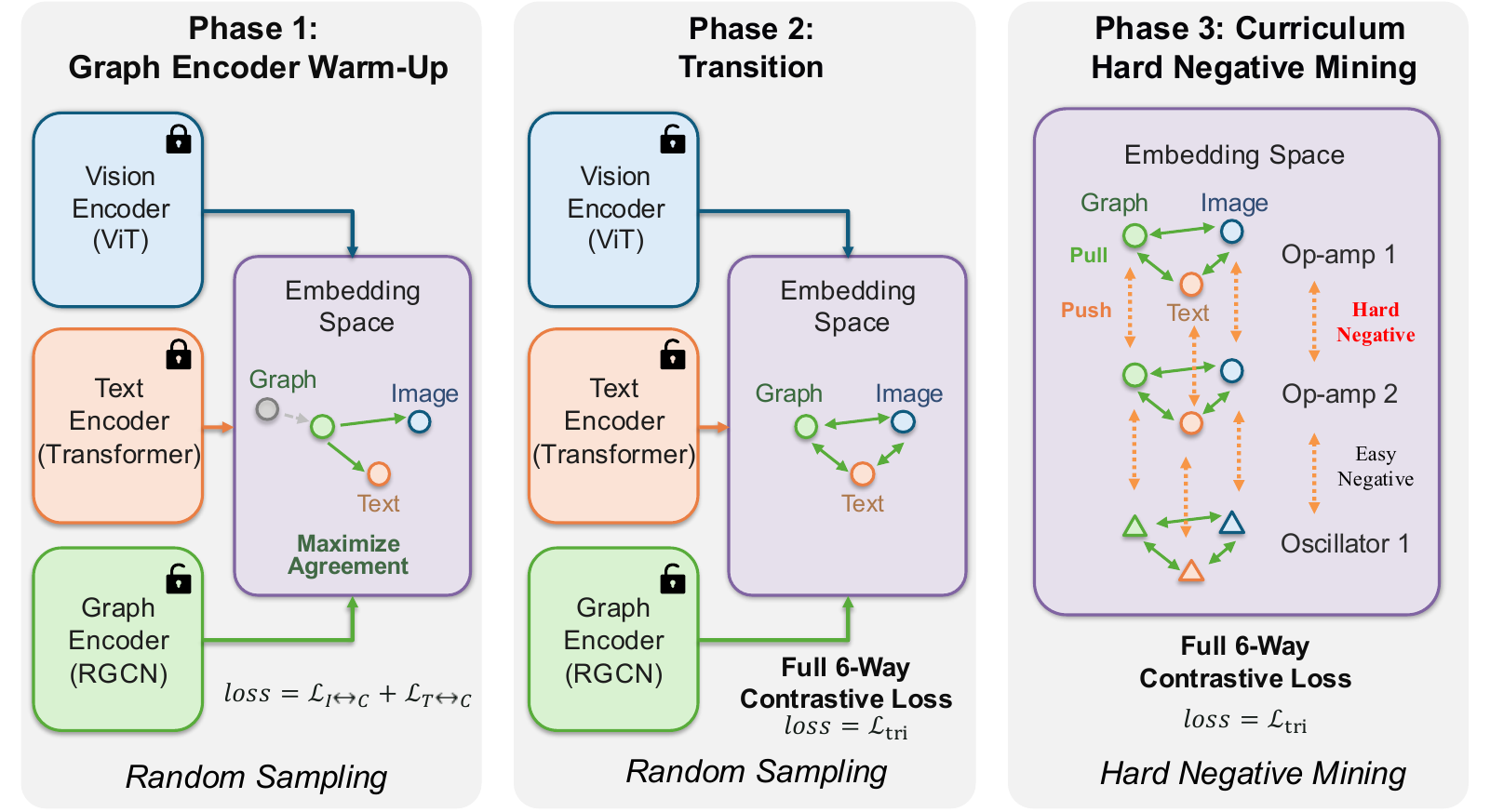}}
    \caption{Overview of the three-phase curriculum training strategy.}
    \label{fig:curriculum}
\end{figure}

Jointly training a randomly initialized RGCN with pretrained CLIP is
unstable: low-quality graph embeddings disrupt the pretrained
alignment, and hard negatives amplify this instability before
cross-modal correspondence is even established. We address this with
the three-phase curriculum illustrated in~\Cref{fig:curriculum},
which progressively increases both the number of trainable parameters
(RGCN~$\to$~RGCN+CLIP) and the sample difficulty
($\alpha_0{=}0.05$~$\to$~$\alpha_{\max}{=}0.3$) as training proceeds. The figure also makes
explicit which loss is active in each phase: Phase~1 uses only the
code-involved losses $\mathcal{L}_{I\leftrightarrow C}{+}\mathcal{L}_{T\leftrightarrow C}$,
while Phases~2 and~3 use the full 6-way tri-modal loss
$\mathcal{L}_{\text{tri}}$. Phase~1 and the switch to
$\mathcal{L}_{\text{tri}}$ are shared by all tri-modal variants in
\Cref{tab:ablation}; the Phase-2 restart and Phase-3 hard negatives are
AnalogRetriever-specific.

\paragraph{Phase~1: Graph Encoder Warm-Up.}
Only the RGCN is trained; both CLIP encoders are frozen. The
alignment loss uses only the code-involved directions,
$\mathcal{L}_{I\leftrightarrow C}{+}\mathcal{L}_{T\leftrightarrow C}$,
with $\alpha{=}0$ (no intra-cluster negatives; see Phase~3). This lets the graph encoder
align to the existing CLIP space without perturbing its pretrained
structure.

\paragraph{Phase~2: Transition.}
We unfreeze CLIP and switch to the full 6-way loss
$\mathcal{L}_{\text{tri}}$ while keeping $\alpha{=}0$, so the model
is not simultaneously asked to handle newly unfrozen parameters
\emph{and} hard negatives. The optimizer is rebuilt with fresh Adam
state and an independent linear-warmup-then-cosine-decay schedule,
establishing a stable joint-optimisation trajectory before hard
negatives are introduced.

\paragraph{Phase~3: Curriculum Hard Negative Mining.}
We cluster all circuits by embeddings of their SPICE netlists, obtained
from Qwen3-Embedding, using $K$-means ($K{=}30$) into
functional clusters
$\{\mathcal{G}_1,\ldots,\mathcal{G}_K\}$. Each batch is built around a
seed circuit drawn uniformly from the training set: an $\alpha_m$
fraction of the batch is drawn from the seed's cluster $\mathcal{G}_k$
and the remainder uniformly from the other clusters. Since the
contrastive loss treats every batch item as a positive, the
intra-cluster items also act as hard negatives for one another. The
ratio increases linearly:
\begin{equation}
    \alpha_m = \min\!\big(\alpha_{\max},\; \alpha_0 + \tfrac{m-1}{M-1} (\alpha_{\max} - \alpha_0)\big),
\end{equation}
where $m$ is the current training epoch within Phase~3, $M$ is the
total number of Phase-3 epochs, $\alpha_0{=}0.05$ is the initial
hard-negative ratio, and $\alpha_{\max}{=}0.3$ is the maximum ratio.
This schedule first
consolidates coarse distinctions (amplifier vs.\ oscillator) and then
progressively sharpens discrimination among structurally distinct
circuits serving similar functions, such as common-source vs.\
common-gate amplifiers or Miller-compensated vs.\ folded-cascode
op-amps.

\section{Experiments}
\label{sec:experiments}
We describe dataset curation, training setup and baselines, main
retrieval results with ablations, and RAG integration into
AnalogCoder.

\subsection{Dataset Curation}
\label{sec:dataset_curation}
We build upon Masala-CHAI~\cite{Bhandari2024MasalaCHAIAL}, the largest
publicly available tri-modal analog circuit dataset (${\approx}6{,}500$
triplets). Our audit revealed severe quality issues: of the 6{,}371
schematic images, only 6{,}069 have paired SPICE netlists and captions,
and among those only 22.0\% compile under Ngspice and a
mere 11.4\% pass a DC operating-point (\texttt{.op}) check
(Table~\ref{tab:data_quality}). Common failure modes include undefined
subcircuits, missing device models, and incorrect node connectivity,
and many captions are generic boilerplate (\textit{``This is a circuit''})
that does not reflect the actual topology.

\begin{figure}[t]
    \centering
    \makebox[\columnwidth][c]{\includegraphics[width=0.96\columnwidth]{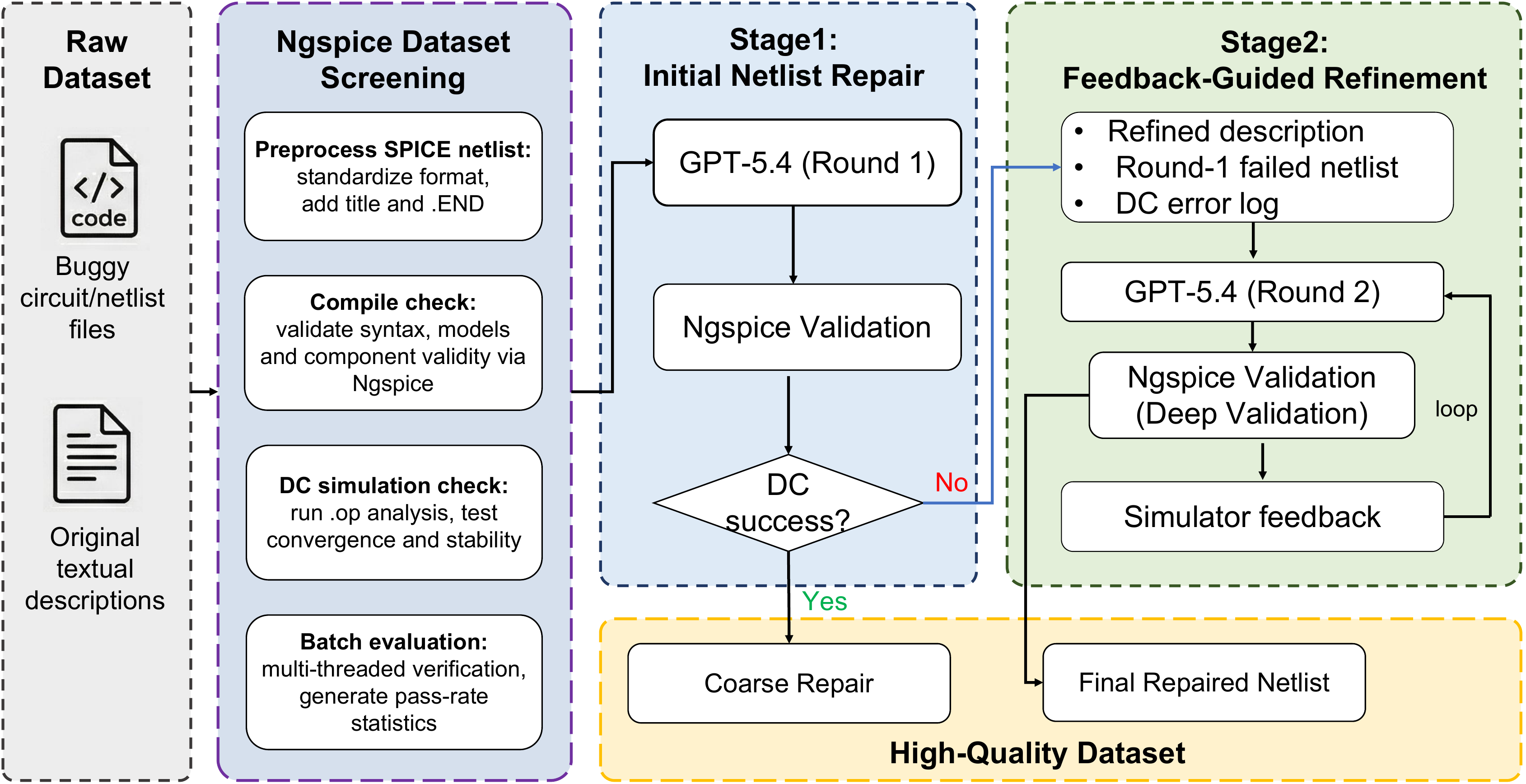}}
    \caption{Overview of the two-stage dataset refinement~pipeline.}
    \label{fig:repair_pipeline}
\end{figure}

\begin{figure}[!t]
\centering
\begin{tcolorbox}[sysprompt, title={Stage~2a: Netlist Repair}]
You are an expert SPICE engineer debugging a netlist that failed Ngspice
compilation or DC simulation. Given the schematic image, the broken
netlist, and the specific Ngspice error messages, return a corrected,
compilable netlist that
\begin{itemize}\setlength\itemsep{1pt}
  \item resolves all reported syntax and reference errors,
  \item provides \texttt{.model} statements for every semiconductor device,
  \item preserves the original topology and function,
  \item includes an \texttt{.op} analysis and ends with \texttt{.END}.
\end{itemize}
Output format: \texttt{<NETLIST>}$\,\ldots$\texttt{</NETLIST>}.
\end{tcolorbox}

\vspace{3pt}

\begin{tcolorbox}[sysprompt, title={Stage~2b: Description Refinement}]
You are an expert analog circuit engineer. Using the verified SPICE
netlist as ground truth (together with the original, possibly
inaccurate caption), produce \emph{one} technically precise sentence
(\,$\le$60 words, max 77 CLIP tokens) that names the circuit type,
lists the key components and their roles, and describes the primary
operating principle.
Output format: \texttt{<DESCRIPTION>}$\,\ldots$\texttt{</DESCRIPTION>}.
\end{tcolorbox}
\caption{Stage~2 prompts: feedback-guided netlist repair and
description refinement.}
\label{fig:stage2_prompt}
\end{figure}

Our refinement pipeline shown in~\Cref{fig:repair_pipeline} has two
cascaded stages connected by an Ngspice simulator acting as the
ground-truth oracle.

\begin{figure}[t]
    \centering
    \makebox[\columnwidth][c]{\includegraphics[width=0.95\columnwidth]{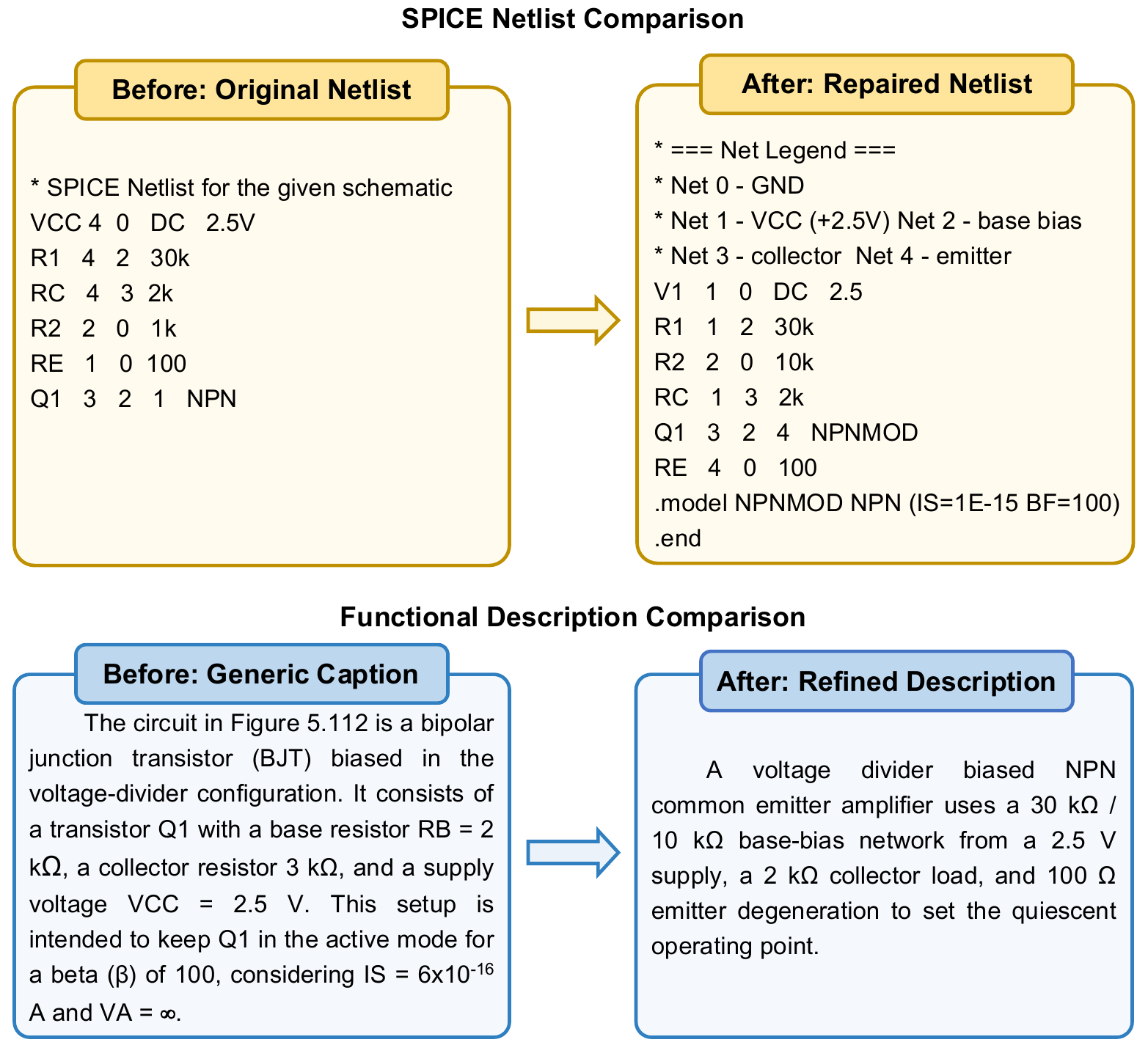}}
    \caption{Before/after refinement. \textbf{Top}: SPICE netlist with
    errors  and fixes. \textbf{Bottom}: generic vs.\ refined
    functional description.}
    \label{fig:data_quality}
\end{figure}

\paragraph{Stage~1: Initial Netlist Repair.}
As shown in the middle block of~\Cref{fig:repair_pipeline},
each raw netlist is fed to GPT-5.4~\cite{gpt54} together with its
Ngspice errors.
Repaired netlists that compile \emph{and} pass the DC check are
committed to the high-quality set; DC failures are forwarded to
Stage~2 with the error log. For the 302 samples originally missing a
netlist or caption, the LLM synthesises both modalities from the
schematic image, recovering all 6{,}371 triplets. Stage~1 alone
raises the compile rate from 22.0\% to 99.2\% and DC pass from
11.4\% to 74.1\%.
\paragraph{Stage~2: Feedback-Guided Refinement.}
For Stage-1 DC failures, we use the two prompts in~\Cref{fig:stage2_prompt}: a \emph{netlist-repair} prompt
(Stage~2a) that fixes the broken netlist from the exact Ngspice error
messages while preserving topology, and a
\emph{description-refinement} prompt (Stage~2b) that rewrites the
caption from the now-verified netlist. At each iteration the model receives the failed netlist and DC error
log and generates a new candidate verified by Ngspice; updated
feedback is appended on failure, for up to $K_{\max}{=}5$ iterations.
Stage~2 lifts the compile rate to 99.97\% and the DC pass rate to
99.7\% (Table~\ref{tab:data_quality}), ensuring text--circuit
consistency. A before/after example appears in~\Cref{fig:data_quality}.

\subsection{Experimental Settings}
\label{sec:settings}

\paragraph{Preprocessing and Training.}
We split all 6{,}371 triplets at the instance level with a seeded
random shuffle (seed 42), hold out the first 1{,}000 for testing, and
train on the remaining 5{,}371. Functional clusters
for the curriculum sampler are obtained by $K$-means ($K{=}30$) on
Qwen3-Embedding representations of the SPICE netlists.
We use CLIP ViT-L/14~\cite{pmlr-v139-radford21a} (bottom 16 of 24 blocks
frozen), a two-layer RGCN~\cite{10.1007/978-3-319-93417-4_38} with hidden dim 512 and
attention pooling, and a two-layer MLP into the 768-d CLIP space.
Training follows the three-phase curriculum of \Cref{sec:curriculum}
with 6, 2, and 12 epochs per phase; $\alpha$ ramps from 0.05 to 0.3
in Phase~3. We use
AdamW (CLIP $5{\times}10^{-5}$, RGCN $5{\times}10^{-4}$), rebuilt at
Phase~2 with an independent warmup, effective batch size 256, and a
learnable logit scale ($1/\tau$) initialised to $1/0.07$. Checkpoints are selected by held-out R@1 (mean of I$\to$C, T$\to$I,
and T$\to$C; T$\to$I for TI). All experiments run on a single NVIDIA
H200 GPU.

\begin{table}[t]
\centering
\caption{Dataset quality before and after refinement.}
\label{tab:data_quality}
\resizebox{\columnwidth}{!}{%
\begin{tabular}{lccc}
\toprule
\textbf{Stage} & \textbf{\# Triplets} & \textbf{Compile (\%)} & \textbf{DC Pass (\%)} \\
\midrule
Original Masala-CHAI              & 6{,}069 & 22.0  & 11.4  \\
+ Stage 1 (Initial Repair)       & 6{,}371 & 99.2  & 74.1  \\
+ Stage 2 (Feedback Refinement)  & 6{,}371 & 99.97 & 99.7  \\
\midrule
Passing both checks               & 6{,}354 & 100.0 & 100.0 \\
\bottomrule
\end{tabular}%
}
\end{table}
\paragraph{Metrics and Baselines.}
We report Recall@$K$ ($K{\in}\{1,5,10\}$) for all six cross-modal
directions and the average R@1 over them. Each evaluation draws a
gallery of $N{=}200$ items uniformly from the 1{,}000 held-out
triplets, ranks candidates by cosine similarity, and averages over ten
draws with fixed seeds (random-guess R@1 is $0.5\%$). Exact-pair Recall@$K$ is a
strict lower bound on usefulness, since several library circuits can
satisfy one query. We compare against four
external retrieval baselines, all applied zero-shot without training
on our corpus:
(i)~\textbf{CLIP}~\cite{pmlr-v139-radford21a} (raw SPICE fed to the off-the-shelf
CLIP text encoder, a zero-shot lower bound);
(ii)~\textbf{CROP}~\cite{11240643} (netlists summarized by
Qwen2.5-7B then CLIP-embedded, a training-free
code-as-text baseline);
(iii)~\textbf{ChatLS}~\cite{11132969} (LLM-based structured
netlist representations); and
(iv)~\textbf{NetTAG}~\cite{10.1109/DAC63849.2025.11133349} (per-device text attributes
encoded with Qwen3-Embedding and pooled).
CROP, ChatLS, and NetTAG only replace the \emph{code} representation,
so all four baselines share the off-the-shelf CLIP image--text pathway
and yield identical T$\to$I and I$\to$T recall.
Following the notation of \Cref{sec:formulation} (I: image, T: text,
C: circuit netlist), the internal ablation variants \textbf{TI
(Bi-Modal)}, \textbf{TIC (GCN)}, and \textbf{TIC (RGCN)} share the
20-epoch budget: TI fine-tunes CLIP on image--text pairs only, while
TIC applies the Phase-1 warm-up followed by the six-way loss but
omits the Phase-2 optimizer restart and the Phase-3 hard negatives.

\subsection{Main Results and Ablation}
\label{sec:main_results}

We present all results, including external baselines and internal ablations, in \Cref{tab:ablation} to facilitate direct comparison.

\begin{table*}[t]
\centering
\caption{Cross-modal retrieval on the held-out set (Recall@$K$, \%). Top: zero-shot baselines; bottom: models trained on our corpus. Best per column in \textbf{bold}.}
\label{tab:ablation}
\resizebox{\textwidth}{!}{%
\renewcommand{\arraystretch}{0.92}%
\begin{tabular}{l ccc ccc ccc ccc ccc ccc c}
\toprule
 & \multicolumn{3}{c}{\textbf{I$\to$C}}
 & \multicolumn{3}{c}{\textbf{T$\to$I}}
 & \multicolumn{3}{c}{\textbf{T$\to$C}}
 & \multicolumn{3}{c}{\textbf{C$\to$I}}
 & \multicolumn{3}{c}{\textbf{I$\to$T}}
 & \multicolumn{3}{c}{\textbf{C$\to$T}}
 & \textbf{Avg} \\
\cmidrule(lr){2-4} \cmidrule(lr){5-7} \cmidrule(lr){8-10}
\cmidrule(lr){11-13} \cmidrule(lr){14-16} \cmidrule(lr){17-19}
\textbf{Model}
 & R@1 & R@5 & R@10
 & R@1 & R@5 & R@10
 & R@1 & R@5 & R@10
 & R@1 & R@5 & R@10
 & R@1 & R@5 & R@10
 & R@1 & R@5 & R@10
 & \textbf{R@1} \\
\midrule
CLIP~\cite{pmlr-v139-radford21a}
 & 1.2 & 4.6 & 8.9
 & 2.2 & 8.2 & 13.5
 & 2.9 & 9.7 & 16.2
 & 2.9 & 10.1 & 15.7
 & 2.7 & 7.6 & 12.9
 & 3.2 & 11.2 & 17.4
 & 2.5 \\
CROP~\cite{11240643}
 & 2.3 & 8.4 & 15.0
 & 2.2 & 8.2 & 13.5
 & 9.2 & 25.8 & 37.1
 & 2.4 & 8.5 & 14.2
 & 2.7 & 7.6 & 12.9
 & 9.4 & 24.7 & 33.0
 & 4.7 \\
ChatLS~\cite{11132969}
 & 0.9 & 5.9 & 11.3
 & 2.2 & 8.2 & 13.5
 & 9.5 & 24.0 & 35.1
 & 1.6 & 6.6 & 10.0
 & 2.7 & 7.6 & 12.9
 & 7.2 & 20.4 & 31.3
 & 4.0 \\
NetTAG~\cite{10.1109/DAC63849.2025.11133349}
 & 0.2 & 2.5 & 4.5
 & 2.2 & 8.2 & 13.5
 & 0.5 & 3.4 & 6.8
 & 0.1 & 1.4 & 4.1
 & 2.7 & 7.6 & 12.9
 & 0.4 & 3.4 & 7.1
 & 1.0 \\
\midrule
TI (Bi-Modal)
 & -- & -- & --
 & 70.5 & 92.9 & 96.8
 & -- & -- & --
 & -- & -- & --
 & 69.8 & 94.1 & 96.7
 & -- & -- & --
 & -- \\
TIC (GCN)
 & 62.9 & 92.2 & 95.7
 & 70.8 & 95.7 & 98.4
 & 67.7 & 95.7 & 99.0
 & 60.9 & 91.6 & 95.6
 & 71.3 & 96.4 & 98.8
 & 65.5 & 95.7 & 98.9
 & 66.5 \\
TIC (RGCN)
 & 65.2 & 92.5 & 96.8
 & 71.8 & 96.2 & \textbf{98.8}
 & 69.1 & \textbf{96.4} & \textbf{99.2}
 & 61.9 & 92.9 & 96.7
 & 71.1 & \textbf{96.7} & \textbf{98.9}
 & 67.3 & 96.4 & \textbf{99.0}
 & 67.7 \\
\rowcolor{blue!5}
\textbf{AnalogRetriever}
 & \textbf{74.7} & \textbf{95.5} & \textbf{98.1}
 & \textbf{78.2} & \textbf{96.8} & 97.9
 & \textbf{75.6} & 96.3 & 98.6
 & \textbf{72.0} & \textbf{94.8} & \textbf{97.9}
 & \textbf{78.5} & \textbf{96.7} & 98.3
 & \textbf{72.4} & \textbf{96.7} & 98.3
 & \textbf{75.2} \\
\bottomrule
\end{tabular}
}
\end{table*}

\paragraph{Tri-modal training benefits bi-modal retrieval.}
On the shared Image$\leftrightarrow$Text directions, the full tri-modal
procedure lifts T$\to$I R@1 from 70.5\% to \textbf{78.2\%} (+7.7) and
I$\to$T from 69.8\% to \textbf{78.5\%} (+8.7) over the bi-modal TI
baseline, despite using the same CLIP backbone and identical
image--text data. Notably, adding the graph encoder alone (TIC vs.\
TI) barely moves T2I ($70.5{\to}70.8$): even with the graph encoder
warmed up against frozen CLIP, joint training without hard negatives
does not by itself sharpen the image--text space. The bi-modal gains
appear only once the Phase-2 restart and the Phase-3 hard negatives
are added.

\paragraph{Curriculum hard-negative mining yields consistent gains.}
Since TIC\,(RGCN) already includes the Phase-1 warm-up and the six-way
loss, comparing it with the full AnalogRetriever isolates the Phase-2
restart and the Phase-3 hard-negative curriculum, which together yield
uniform improvements across all six directions ($+9.5$ on I$\to$C,
$+10.1$ on C$\to$I, $+6.4$ on T$\to$I, $+7.4$ on I$\to$T, $+6.5$ on
T$\to$C, and $+5.1$ on C$\to$T). The largest gains appear on
\emph{code-involved} directions, where the hard-negative sampler
forces the model to distinguish circuits that share similar
high-level descriptions but differ in transistor-level topology
(e.g., a folded-cascode op-amp versus a two-stage Miller op-amp).
Overall, average R@1 improves from 67.7\% to \textbf{75.2\%} (+7.5).
Port-aware RGCN adds $+1.2$ Avg R@1 over the edge-agnostic GCN by
encoding device-port semantics, with the gains concentrated on the
code-involved directions. Every AnalogRetriever direction
exceeds 94\% at R@5 and 97\% at R@10, so a user query almost always
places the correct circuit within the top-10 candidates.

\subsection{Retrieval-Augmented Generation}
\label{sec:rag}

\begin{table}[!tbp]
  \centering
  \caption{Functional correctness (\%) on the AnalogCoder benchmark
  (24 tasks $\times$ 5 trials) for eight LLMs without and with
  AnalogRetriever; $\Delta$: absolute gain.}
  \label{tab:rag_cross_model}
  \begin{tabular}{lccc}
    \toprule
    \textbf{LLM} & \textbf{Baseline} & \textbf{+RAG} & \textbf{$\Delta$} \\
    \midrule
    GPT-4o-mini        & 30.8 & 40.8 & \textbf{+10.0} \\
    Nemotron-120B      & 34.2 & 37.5 & +3.3 \\
    GPT-5.4-mini       & 65.0 & 67.5 & +2.5 \\
    Gemini-3-Flash     & 65.0 & 70.0 & +5.0 \\
    Kimi-K2            & 65.0 & 73.3 & +8.3 \\
    GLM-4.6            & 73.3 & 80.0 & +6.7 \\
    Qwen3.5-397B       & 78.3 & 85.0 & +6.7 \\
    Claude-Sonnet-4.6  & 84.2 & \textbf{86.7} & +2.5 \\
    \midrule
    \textbf{Average}   & \textbf{62.0} & \textbf{67.6} & \textbf{+5.6} \\
    \bottomrule
  \end{tabular}
\end{table}

To demonstrate practical utility, we integrate AnalogRetriever into a
RAG pipeline with
AnalogCoder~\cite{Lai_2025}, a training-free LLM agent for
analog design via PySpice code~generation.

\paragraph{RAG Pipeline.}
AnalogRetriever ranks the training-set netlists by cosine similarity to
the encoded query via Text$\to$Code search, and a device-family gate keeps
the highest-ranked candidates whose dominant device family matches the
family inferred from the task description. The selected netlists are
converted to PySpice by a rule-based parser \emph{before} prompt injection,
so in-context examples use the same API as the target code.

\paragraph{Cross-Model Evaluation.}
We evaluate on AnalogCoder's 24-task benchmark (5~trials $\times$
3~retries) across eight LLMs (Table~\ref{tab:rag_cross_model}).
AnalogRetriever delivers a positive gain on \emph{all eight}
models, averaging \textbf{+5.6\%} absolute
(62.0\%$\to$67.6\%). The largest gain is on GPT-4o-mini (+10.0);
augmenting Claude~Sonnet~4.6 reaches \textbf{86.7\%}, a new state
of the art. The benefit \emph{generalizes across model families and
scales}.

\begin{figure*}[!tbp]
      \centering
      \includegraphics[width=0.90\textwidth]{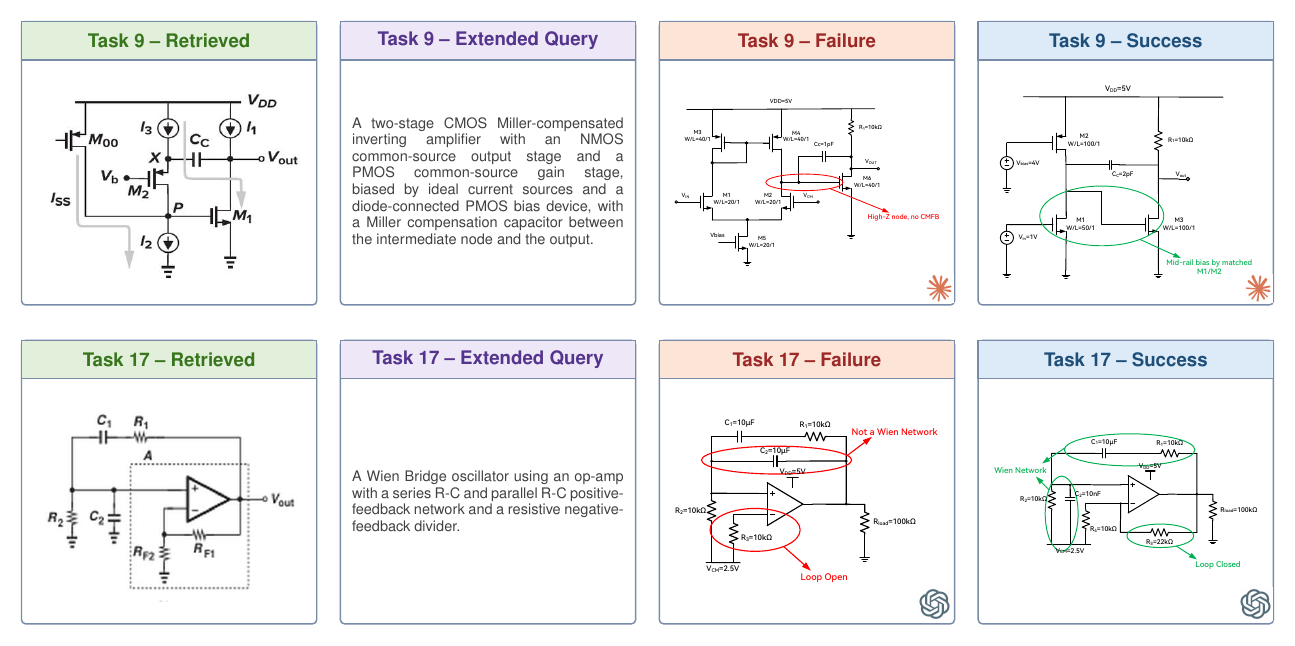}
     \caption{Case studies of retrieval-augmented circuit generation on Tasks 9 and 17.}
      \label{fig:case_study}
\end{figure*}

\paragraph{Qualitative Case Studies.}
\Cref{fig:case_study} shows two representative cases.
In Task~9 (top row), Claude~Sonnet~4.6 without RAG produces a diff-pair 
with a current-mirror load but no common-mode feedback, letting the 
high-impedance node drift into triode ($0/5$). Grounded by a retrieved 
single-ended CMOS Miller op-amp, the LLM reproduces the canonical 
M1$\to$M3 topology with a Miller capacitor, reaching $5/5$.
In Task~17 (bottom row), GPT-5.4-mini without retrieval produces an 
incorrect RC network structure and leaves the feedback loop open, 
preventing oscillation ($0/5$). With a retrieved reference, the LLM 
constructs the correct Wien-bridge topology with a closed feedback 
loop, achieving $4/5$ success. 
Both cases demonstrate that retrieval provides the topological 
guidance needed to turn invalid outputs into functional circuits.

\paragraph{Effect of Query Expansion.}
AnalogCoder task prompts are typically only 3--6 tokens long (e.g.,
\emph{``A Wien Bridge oscillator''}), which CLIP's text encoder maps
to low-discriminative embeddings. Rewriting each task into a
topology-aware specification lifts relevant database entries from deep
ranks to the top-$k$ (e.g., the canonical Wien-bridge op-amp entries
move from ranks 74/56/20 to 2/10/5; the Miller op-amp from rank~35 to
rank~1). We therefore let GPT-5.4 rewrite every task once into a
topology-aware paragraph before retrieval; the cost is amortized
across all 24 tasks and negligible. \Cref{fig:case_study}
visualizes both the extended query and its top-1 reference for each
case study; all RAG results use expanded queries.


\section{Conclusion}
\label{sec:conclusion}
We presented AnalogRetriever, the first tri-modal retrieval
model that aligns natural-language descriptions, schematic images, and
SPICE netlists in a shared embedding space, supported by a two-stage
LLM-based refinement pipeline that lifts the Masala-CHAI DC pass rate
from $11.4\%$ to $99.7\%$ and yields $6{,}371$ refined triplets.
A port-aware RGCN encoder trained with a three-phase curriculum and
hard-negative mining pushes average R@1 to $\mathbf{75.2\%}$
across all six cross-modal directions, outperforming the best zero-shot
prior baseline by over an order of magnitude. Plugged into
AnalogCoder, AnalogRetriever delivers a positive
functional-correctness gain on \emph{all eight} evaluated LLMs
(averaging $+5.6\%$ absolute) and sets a new state of the art of
$\mathbf{86.7\%}$ on Claude~Sonnet~4.6, with the benefit generalizing
across model families and parameter scales rather than being tied to
any specific LLM.

Three findings from our experiments deserve particular emphasis.
First, \emph{tri-modal training yields mutual enhancement}: even the
Image$\leftrightarrow$Text directions, which share the same CLIP
backbone and identical training data, improve by up to $+8.7$ R@1 over
the bi-modal baseline under the full tri-modal procedure, showing that
the graph encoder and the hard-negative curriculum together enrich the
shared embedding space in ways that benefit all modalities, not just
the newly introduced one.
Second, \emph{the hard-negative curriculum, not the graph encoder
alone, drives the gain}: a tri-modal model that already warms up the
graph encoder and trains the full six-way loss still trails
AnalogRetriever by $+7.5$ Avg R@1, which the Phase-2 optimizer restart
and the Phase-3 curriculum recover.
Third, our qualitative case studies show that
a single high-quality retrieved reference is often enough to
re-anchor a generation that would otherwise commit to a topologically
invalid circuit family, turning complete failures into reliable
successes, suggesting that retrieval and generation are complementary
rather than competing paradigms for analog design automation.

\paragraph{Limitations and future work.}
Several directions remain open.
(i)~The current dataset covers 19 canonical analog topologies; extending 
to mixed-signal, RF, and power-management circuits would broaden 
applicability.
(ii)~The RGCN uses 19 hand-defined port types; learning the relation 
vocabulary from data could improve generalization to novel device 
technologies.
(iii)~As circuit databases grow to industrial scale, efficient 
nearest-neighbor search (e.g., product quantization) becomes essential.
We will release the curated dataset and model weights.


\balance
\begin{acks}
This work was supported by the National Key Laboratory of Integrated
Circuits and Microsystems (No.~NICL2026KF1004) and the Natural Science
Research Start-up Foundation of Recruiting Talents of Nanjing University
of Posts and Telecommunications (Grant No.~NY225066).
\end{acks}

\clearpage
\apptocmd{\thebibliography}{\renewcommand{\newblock}{}}{}{}
\bibliographystyle{ACM-Reference-Format}
\bibliography{ref}

\end{document}